\documentclass[letterpaper, 10 pt, conference]{ieeeconf}  

\IEEEoverridecommandlockouts                              

\usepackage{graphicx}
\usepackage{float}
\usepackage{subfig}
\usepackage{amsfonts}
\usepackage{bm}
\usepackage{hyperref}
\newcommand{\etal}{{et al}.\@ } 

\title{TridentNetV2: Lightweight
  Graphical Global Plan Representations for Dynamic Trajectory
  Generation}

\author{David Paz$^{1}$, Hao Xiang$^{1}$, Andrew Liang$^{1}$, and Henrik I. Christensen$^{1}$
  \thanks{$^{1}$Contextual Robotics Institute, University of
    California San Diego, La Jolla, CA 92093, USA {}}%
}

\begin{document}

\maketitle
\thispagestyle{empty}
\pagestyle{empty}

\begin{abstract}
  We present a framework for dynamic trajectory generation for
  autonomous navigation, which does not rely on HD maps as the
  underlying representation.  High Definition (HD) maps have become a
  key component in most autonomous driving frameworks, which include
  complete road network information annotated at a centimeter-level
  that include traversable waypoints, lane information, and traffic
  signals.  Instead, the presented approach models the distributions of
  feasible ego-centric trajectories in real-time given a nominal graph-based
  global plan and a lightweight scene representation.  By embedding
  contextual information, such as crosswalks, stop signs, and traffic signals, our
  approach achieves low errors across multiple urban navigation
  datasets that include diverse intersection maneuvers, while maintaining
  real-time performance and reducing network complexity. Underlying
  datasets introduced are available online.
\end{abstract}

\section{INTRODUCTION}

In autonomous navigation, a global plan defines the sequences of
instructions needed for point-to-point navigation. These plans can be
generated using various types of maps, including High Definition (HD)
and Coarse maps. HD maps (e.g.\ nuScenes \cite{caesar2020nuscenes},
Waymo Open Dataset \cite{sun2020scalability}, Argoverse
\cite{chang2019argoverse}, etc.) are widely used today by recent
autonomous driving architectures; they provide rich contextual
information such as road network definitions, lane markings, crosswalks, and drivable areas. Given that HD maps include complete road networks and lane definitions, trajectory information is
readily available during path planning. This facilitates the process of identifying feasible trajectories for navigation, path tracking,
and control. Nevertheless, various limitations arise in dynamic
environments such as construction sites or road changes. If the environment drastically changes between the time the map was last
updated and the time the autonomous agent is navigating, the agent may
be unable to operate using the outdated map. A second constraint
involves scalability for large scale deployments; today, most maps are
labeled manually or only partially automated. 

\begin{figure}
  \begin{minipage}{.49\linewidth}
    \centering
    \subfloat[]{\label{figure:1a}\includegraphics[trim={20cm 4cm 20cm 3.55cm}, clip, width=\textwidth]{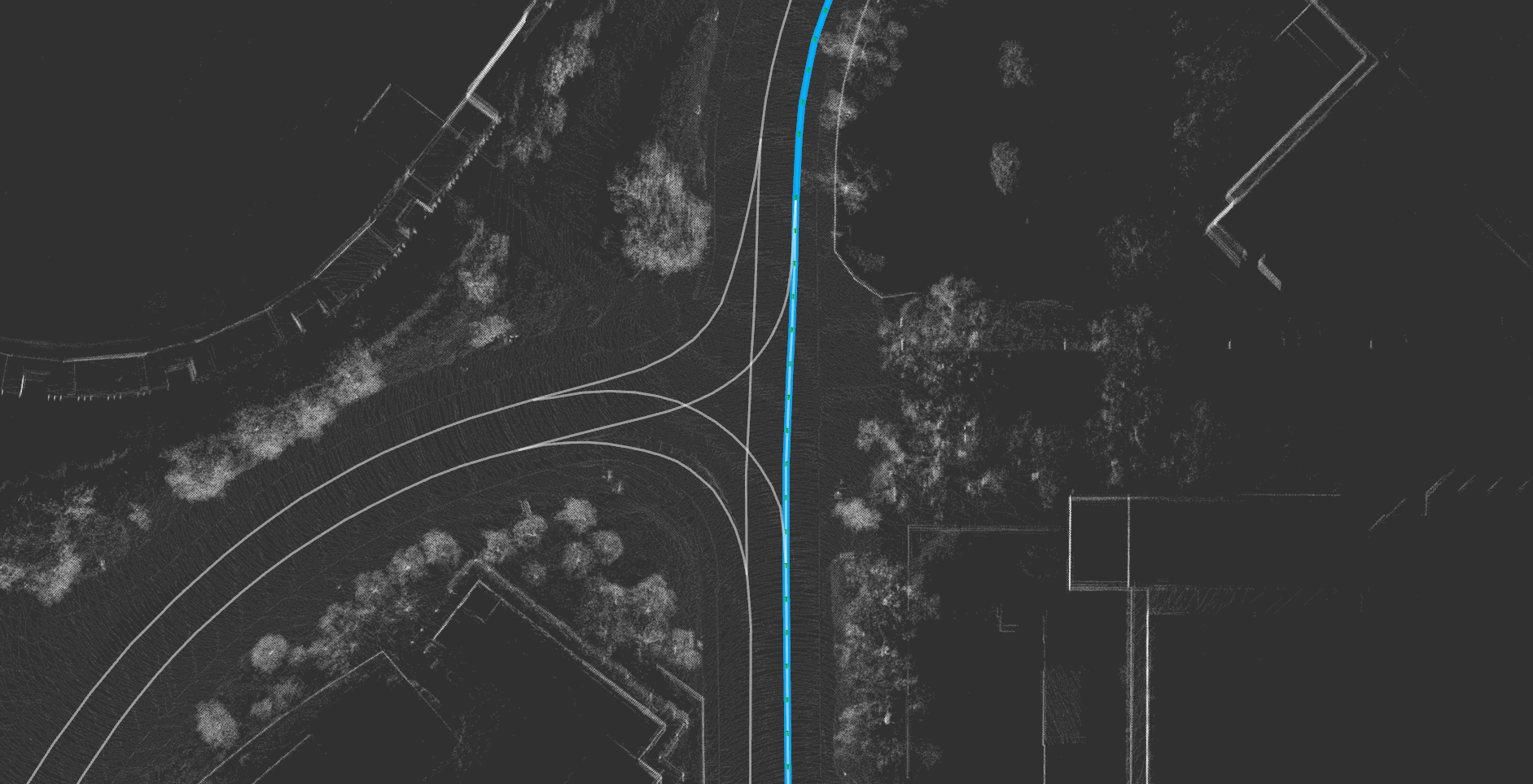}}
  \end{minipage}%
  \hfill
  \begin{minipage}{.49\linewidth}
    \centering
    \subfloat[]{\label{figure:1b}\includegraphics[width=\textwidth]{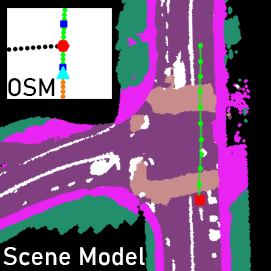}}
  \end{minipage}%
  \par\medskip
  
  \caption{Illustrations of global planning and scene representations
    for autonomous navigation: (a) trajectory generated using HD maps
    (blue trajectory represents plan/gray trajectories represent
    complete roadnetwork), (b) a trajectory generated dynamically (shown as the green
    trajectory)
    using a nominal OSM global plan and an
    automatically generated semantic scene representation without HD maps. }
  \label{figure:1}
\end{figure}

To this end, a coarse map representation  for
global planning can present advantages over HD maps in dynamic environments due to scalability
implications; examples of coarse maps include open-source maps (i.e. OpenStreetMaps)
and proprietary maps (i.e. Google Maps). These lightweight maps often
include labels to describe road segments and intersections; however,
lane and trajectory information is not included. This
presents an open research question on identifying traversable
trajectories dynamically given a scene representation.

Most recently, this dynamic trajectory generation task was formulated
by the TridentNet Conditional Generative Model (CGM)
\cite{paz2021tridentnet} using rasterized global plan and semantic
scene representations. The approach achieves low relative errors on an
urban dataset in real-world scenarios
without use of HD maps. Our work extends this formulation by
introducing TridentNetV2; a model that utilizes graphical global plan
representations instead of rasterized representations. The graphical
global plans introduced are encoded from map data extracted from
OSM that includes information about road segments,
crosswalks, traffic signals, and stop signs.  A Self-Attention
mechanism \cite{10.5555/3295222.3295349} is then applied to generate a
global plan embedding that is then combined with a semantic scene
representation. Finally, a Conditional Variational Autoencoder (CVAE)
\cite{NIPS2015_8d55a249} is leveraged to regress feasible ego-centric
trajectories that account for the multimodal nature of urban
navigation. Our key contributions are as follows:

\begin{itemize}

 \item We introduce a lightweight graph-based conditional generative model that can estimate ego-centric trajectories in real-time for the urban navigation task. In contrast to recent developments, our approach is formulated without relying on manually labeled HD maps.
 \item We further evaluate our formulation towards specific tasks by introducing a new dataset, IntersectionScenes1.0, specifically for intersection navigation. The new open-source dataset is publicly available.\footnote{Datasets: avl.ucsd.edu}
 \item A quantitative comparison is performed with respect to a baseline model across two datasets; the approach shows improvements in terms of error reduction while reducing the model complexity by 31\%.

\end{itemize}

\section{RELATED WORK}

\subsection{HD maps for Autonomous Navigation}

Various methodologies for autonomous navigation in urban environments
have been introduced over the years. Examples of classical motion
planning and control strategies that utilize HD maps include Autoware
\cite{7368032} and Apollo \cite{apollo17}. These frameworks define
clear layers of abstraction for localization, perception, planning,
and control modules that facilitate integration with HD maps; an
example of a plan generated using HD maps is shown in
Fig. \ref{figure:1a}. Learning-based strategies have also been
introduced along HD maps that bypass the layers of abstraction: examples include imitation
learning \cite{bansal2018chauffeurnet}, methods with intermediate
interpretable representations \cite{zeng2019end}, and mixtures of
affine time-varying systems \cite{ivanovic2021mats}. Furthermore, various mapping and road element estimation methods~\cite{paz2020probabilistic,zhang2021hierarchical,xu2021holistic} have been proposed to extract environment context dynamically and reduce the reliance of expensive HD maps. In this work, we employ a probabilistic semantic mapping method~\cite{paz2020probabilistic} to provide additional semantic context besides the coarse road features provided by OSM.

\subsection{Nominal maps and End-to-End models}

Given that HD maps require high computational overhead, manual
labeling, and continuous updates, methods that leverage nominal maps
for navigation are of paramount interest. For this reason, learning-based strategies have utilized lightweight global plan representations
to learn the distribution of potential steering control actions that
an autonomous vehicle can take. For example,
\cite{amini2019variational} combined rasterized OSM global plans with
raw image data to learn control actions with a variational
autoencoder. Similarly, \cite{hecker2020learning} introduced a
generative adversarial network to learn the control inputs using raw
image data. Although these methods would ideally operate in unseen
environments, they can suffer of severe compound errors associated
with imitation learning and are trained with vehicle and sensor
specific configurations. Moreover, as discussed in
\cite{gao2020vectornet}, the image-based methods are more
computationally expensive. Finally, the sensor and vehicle specific
representations leveraged as input features may not be applicable for
new robots and different sensor configurations.

\subsection{Dynamic Trajectory Generation}
A recent research direction involves extracting visual cues from
automatically generated scene representations to estimate ego-centric
trajectories in real-time without depending on HD maps. In addition to
operating with nominal representations, the trajectories generated are
decoupled from motion planning, perception, and control. This can
facilitate integration with arbitrary sensor suites and vehicle
platforms without training a new model. More specifically, TridentNet
CGM \cite{paz2021tridentnet} combines a rasterized global plan
representation with an aggregated probabilistic semantic scene model
\cite{paz2020probabilistic} to generate dynamic trajectories with a
CVAE formulation. Although the semantic map can encode contextual
information, the relationship between objects is not explicitly
modeled. Thus, a compact graphical representation to model object
relationships may be needed. On the other hand, Gao \etal
\cite{gao2017intention} proposed using indoor maps to compute the
high-level path to the goal, which are usually not accessible for
general outdoor environments. Finally, \cite{casas2021mp3} introduces
an approach that unifies perception and planning with interpretable
representations to generate plans via aggregated LiDAR scans, but the approach may not generalize
for complex maneuvers that are not within the predefined driving
commands including ‘keep lane,’ ‘turn left,’ and ‘turn right.’

\section{GRAPHICAL REPRESENTATIONS FOR TRAJECTORY GENERATION}

\begin{figure}
\vspace{1.5mm}
\centering
  \begin{minipage}{.49\linewidth}
    \centering
    \subfloat[]{\label{figure:gp1}\includegraphics[width=\textwidth]{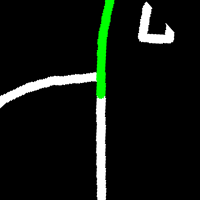}}
  \end{minipage}%
  \hfill
  \begin{minipage}{.49\linewidth}
    \centering
    \subfloat[]{\label{figure:gp2}\includegraphics[width=\textwidth]{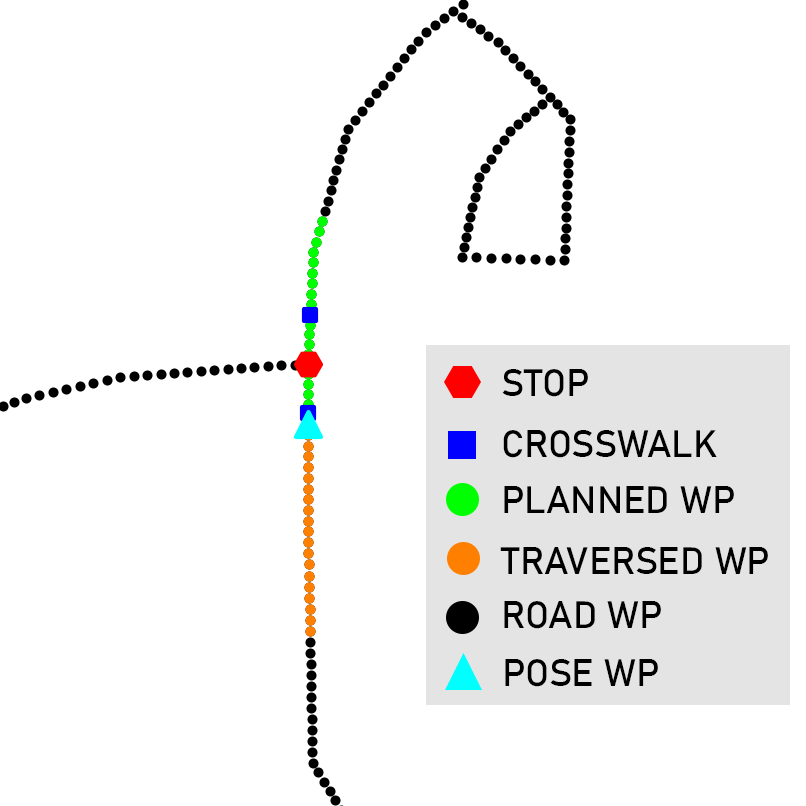}}
  \end{minipage}%
  \par\medskip

  \caption{Nominal representation using OSM, (a) rasterized and (b)
    graphical. These global plans correspond to the scenario defined
    in Fig. \ref{figure:1}}
\label{figure:gp}
\end{figure}

\begin{figure*}
  \centering
  \includegraphics[scale=0.83]{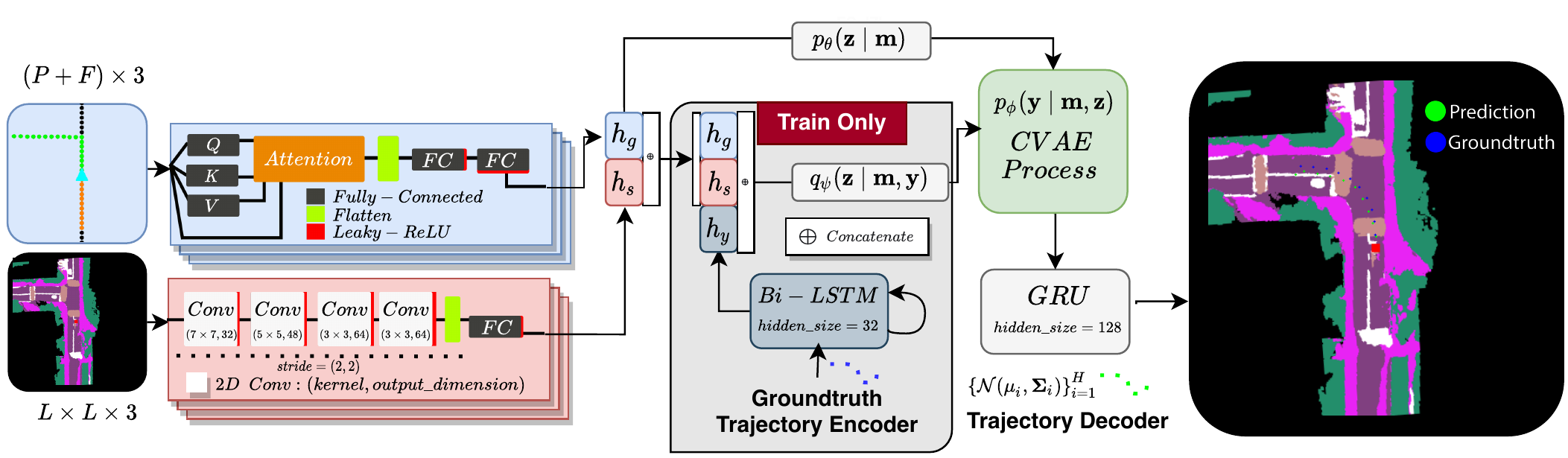}
  \caption{The proposed architecture leverages a Self-Attention
    mechanism to encode the graph-based global plans, a sequence of 2D
    convolutional layers to encode the local scene representation
    around the ego-vehicle, and finally utilizes a CVAE formulation to
    dynamically generate ego-centric trajectories using a GRU. The
    Bi-LSTM is only applied during training phase to characterize
    $q_{\psi}(\mathbf{z} \mid \mathbf{m}, \mathbf{y})$. A visualization of the data flow can be found at \url{https://youtu.be/KCApJMirOC0}}
    \label{fig:architecture}
\end{figure*}

In this section, the method and representations used for TridentNetV2
are introduced. An overview of the model proposed is shown in
Fig. \ref{fig:architecture}. This approach is formulated using a
global plan and automatically generated semantic maps. The global plan
is represented as a graph $\mathbf{m}_g$ which defines road connectivity
and the high-level instructions needed to reach a particular
destination given a GPS pose or starting point. This global plan
representation serves as a coarse directional cue for the trajectory
generation module. On the other hand, a local semantic representation
$\mathbf{m}_s$ describes nearby features such as drivable areas, lane
markings, and sidewalks in the vehicle frame. Both
representations, $\mathbf{m}_g$ and $\mathbf{m}_s$, are updated as
part of the localization process every time a new ego-vehicle pose
update is received. A CVAE approach is then applied during each update
to model the distribution of potential trajectories
$p\left(\mathbf{y} \mid \mathbf{m}\right)$ that can be executed by the
ego-vehicle given the global plan and the semantic scene
representation; where $\mathbf{y}=\{(x_i, y_i)\}^{H}_{i=1}$ is the
trajectory generated dynamically with a horizon $H$ and
$\mathbf{m}= \{ \mathbf{m}_g,\mathbf{m}_s\}$ is a joint embedding for
$\mathbf{m}_g$ and $\mathbf{m}_s$. These features can jointly provide
detailed semantics that encode information about the scene and the
global plan; thus, helping the network learn the relationship between
road elements.

\subsection{Global Plan Representation}
\label{sec:global-plan-representation}


To reason about point-to-point navigation and determine if a left-turn
or a right-turn is needed during intersection navigation, a global
plan is needed. Previously, \cite{Hecker_2018_ECCV,
  amini2019variational, paz2021tridentnet} applied rasterized
representations that were generated using OpenStreetMaps to encode
high-level information. An example can be visualized in
Fig. \ref{figure:gp1}. This image-based global plan encodes the plan
needed to reach a particular destination by utilizing GPS to estimate
a rough pose, an IMU to estimate heading (yaw), and odometry data to
update the rough vehicle state in between GPS measurements. To prevent
bias towards a particular orientation or maneuver type, this global
plan is represented in the ego-vehicle frame by utilizing the heading
to perform a 2D rotation.

To further explore the implications of global plan representations on
dynamic trajectory generation, the original global plan data from
\cite{paz2021tridentnet} is extended with graphical information
(Fig. \ref{figure:gp2}). Instead of rasterizing the global plan
information generated by a GPS-based planner, the plan is directly
treated as a graph with feature vectors that can provide the
flexibility of recent graphical methods. Afterwards, IMU-based integration is used to convert the global plan from the map
frame to an ego-centric frame. In the TridentNet baseline compounding
errors are observed due to IMU-based integration. While correction
procedures can be considered, our approach instead uses the trajectory
information directly provided by Open Street Maps to estimate heading information.
Finally, the graphical representation is augmented with additional
road elements including stop signs, traffic
signals, and pedestrian crossings. For the experiments described in
the next section, each node within the graph $\mathbf{m}_g = (V, E)$
is represented by a three-dimensional vector $(p_x, p_y, f)$ that
corresponds to the node position in meters $(p_x, p_y)$ with a feature
$f$ that represents a unique value depending on if the node is a stop
sign, pedestrian crossing, traffic signal, or along the planned
trajectory and traversed trajectory. To simplify the model, only the
nodes along the traversed trajectory and planned trajectory are
considered and yield a representation with dimensions
$(P+F) \times 3$; where $P$ corresponds to the length of the traversed
trajectory and $F$ is the length of the planned trajectory. In our
experiments, $P=F=20$.

\subsection{Semantic Scene Representation}


Although the global plan approach using OSM can encode the high-level
instructions for reaching a destination, these are low-accuracy and do
not provide additional information to reason about lane markings and
driveable areas. This information is needed for precise path tracking
and navigation.  To incorporate this contextual information, the
trajectory generation model is conditioned on a local semantic scene
representation $\mathbf{m}_s\in \mathbb{R}^{L \times L \times 3}$,
where $L=2D\cdot L_{max}$, $D=2$ corresponds to the discretization
factor in terms of $pixel/m$ and $L_{max}=100m$ is the lateral and
longitudinal horizon for the local semantic map
(Fig. \ref{figure:1b}). First, a 2D semantic map is generated
automatically by driving once along areas of interest and is
post-processed as described in \cite{paz2020probabilistic}; the map is
discretized by a factor $D$ and is represented by an image that
encodes information about driveable areas, sidewalks, crosswalks, lane
markings, and vegetation. Finally, given that contextual information
is only necessary for navigation within a limited horizon, we perform
an $L \times L$ region crop process at run-time by leveraging
localization to perform an ego-centric coordinate transformation. This
local semantic scene representation is then used as input for our
model. An advantage of this approach is that it accounts for road
segments with steep inclines and curved roads by leveraging camera-LiDAR
projective geometry techniques.

\subsection{Dynamic Trajectory Generation using Graphical Global Plans}
A Conditional Variational Autoencoder is a deep conditional generative
model \cite{NIPS2015_8d55a249} that implements a directed graphical
model to approximate a conditional distribution
$p\left(\mathbf{y} \mid \mathbf{m}\right)$. This directed graphical
method explicitly models various distribution modes by introducing a
latent variable $\mathbf{z}$ that is drawn from the prior distribution
$p\left(\mathbf{z} \mid \mathbf{m} \right)$. An advantage of this formulation is that it can capture the multi-modal nature of a distribution, i.e., for navigation, this involves left turns, right turns, u-turns, etc. For discrete latent
variables, $\mathbf{z}$ can be marginalized to recover the original
distribution
$p\left(\mathbf{y} \mid \mathbf{m}\right)=\sum_{\mathbf{z} \in
  \mathbf{Z}} p_{\phi}\left(\mathbf{y} \mid\mathbf{m},
  \mathbf{z}\right) p_{\theta}\left(\mathbf{z} \mid \mathbf{m}
\right)$. As part of the derivation, a \textit{recognition} model,
$q_{\psi}(\mathbf{z} \mid \mathbf{m}, \mathbf{y})$, is introduced
during training to learn a better approximation for
$p_{\theta}\left(\mathbf{z} \mid \mathbf{m} \right)$ since it has
access to groundtruth trajectory data $\mathbf{y}$. The empirical objective
function is then derived as follows
$\left.\mathcal{L}_{cvae}=-\mathbb{E}_{q} [\log
  p_{\phi}\left(\mathbf{y} \mid \mathbf{m}, \mathbf{z}\right)\right]+
\mathbb{KL}\left[q_{\psi}(\mathbf{z} \mid \mathbf{m}, \mathbf{y}) \|
  p_{\theta}(\mathbf{z} \mid \mathbf{m})\right]$.

Motivated by the multimodal capabilities of CVAEs, we extend this
formulation for dynamic trajectory generation for the urban driving
scenario. Our approach follows a similar strategy as derived in
\cite{paz2021tridentnet}; however, we leverage a graphical
representation to encode the global plan. Additionally, we find that
by introducing a mean-squared error (MSE) loss term, lower errors can
be achieved with respect to groundtruth trajectories. Thus the overall
objective function that we seek to minimize is given by
Eq.~\ref{eq:1}, where $\hat{\mathbf{y}}$ corresponds to the predicted
trajectories.

\begin{equation}
  \label{eq:1}
  \mathcal{L}=\mathcal{L}_{cvae}+\frac{1}{H} \sum_{i=1}^{H}\left\|\mathbf{y}_{i}-\hat{\mathbf{y}}_{i}\right\|^{2}
\end{equation}

\subsubsection{Implementation}
Multiple encoder modules are employed to embed the graphical
representation of the global plan and the local semantic scene
representation. To account for
relationships across various road elements and the traversed/planned
trajectories from the global plan, a Self-Attention mechanism
\cite{10.5555/3295222.3295349} is applied within the global plan
encoder. The attention operation is defined by Eq. \ref{eq:2}, where $C=3$ and
$\mathbf{Q}$, $\mathbf{K}$, and $\mathbf{V}$ are linear projections of
$\mathbf{m}_g$ that are referred to as queries, keys, and values,
respectively. A multi-layer perception (MLP) is then applied to the
vectorized representation of the output; thus, producing a global plan
embedding $\mathbf{h}_g$.
\begin{equation}
  \label{eq:2}
  \textit{Attention}(\mathbf{m}_g)=\mathbf{m}_{g}+\textit{softmax}\left(\frac{\mathbf{Q} \mathbf{K}^{T}}{\sqrt{C}}\right) \mathbf{V}
\end{equation}

To encode the $L \times L \times \ 3$ tensor that describes the scene
in a local frame, a sequence of convolutional and fully connected
layers is applied. This semantic scene encoder generates a compressed
representation $\mathbf{h}_s$ of the semantic model.

Lastly, the CVAE is implemented with 12 different modes
($|\mathbf{Z}| = 12$) to encode the multi-modal nature of navigation;
it is assumed that $p_{\theta}(\mathbf{z} \mid \mathbf{m})$ and
$q_{\psi}(\mathbf{z} \mid \mathbf{m}, \mathbf{y})$ are categorical
distributions, where $p_{\theta}(\mathbf{z} \mid \mathbf{m})$ is
parameterized by concatenating $\mathbf{h}_g$ and $\mathbf{h}_s$. On
the other hand, to parameterize
$q_{\psi}(\mathbf{z} \mid \mathbf{m}, \mathbf{y})$, a groundtruth
trajectory feature embedding $\mathbf{h}_y$ is produced by applying a
bi-directional LSTM; $\mathbf{h}_g$
and $\mathbf{h}_s$ are then concatenated with $\mathbf{h}_y$ to
characterize $q_{\psi}(\mathbf{z} \mid \mathbf{m},
\mathbf{y})$. 

During training time, $p_{\theta}(\mathbf{z} \mid \mathbf{m})$ and
$q_{\psi}(\mathbf{z} \mid \mathbf{m}, \mathbf{y})$ are jointly
optimized by following Eq.~\ref{eq:1}. To compute
$\log p_{\phi}\left(\mathbf{y} \mid \mathbf{m}, \mathbf{z}\right)$,
$\mathbf{y}$ is
assumed to be i.i.d and $p_{\phi}\left(\mathbf{y} \mid \mathbf{m}, \mathbf{z}\right)$ is characterized by $H$ bi-variate Gaussian
distributions, namely,
$\left\{\mathcal{N}\left(\bm{\mu}_{i},
    \mathbf{\Sigma}_{i}\right)\right\}_{i=1}^{H}$, where $H$
corresponds to the number of waypoints that define each trajectory and
Gated-Recurrent Unit (GRU)~\cite{cho-etal-2014-learning} is leveraged to generate $\bm{\mu}_i\in \mathbb{R}^2$ and
$\mathbf{\Sigma}_i \in \mathbb{R}^{2\times 2}$ in a recurrent manner.

During testing, only the mode $\mathbf{z}^{*}$ with the highest score
is sampled such that
$\mathbf{z}^{*}=\arg \max _{\mathbf{z}} p_{\theta}(\mathbf{z} \mid
\mathbf{m})$. This mode is then used to decode the predicted
trajectory with the GRU.

\section{Experiments}
The data associated with these experiments is generated through the
calibration, data collection, and semantic mapping process described
in \cite{paz2020probabilistic}. The groundtruth trajectories provided
are automatically annotated using localization. To prevent bias
towards speed, each of the trajectories is interpolated before
training and extends to a 30m horizon that is characterized by $H=10$
waypoints spaced 3m apart. For reference, the global plan encodes
information with a precision range of approximately 1m. On the other
hand, the trajectories annotated in the semantic map frame are
annotated within a 2cm range.

\subsection{Datasets}
To evaluate the adopted global plan representation,
the global plan data (Fig. \ref{figure:gp1}) introduced by
\cite{paz2021tridentnet} (NominalScenes1.0) is extended with its
corresponding graph representation (Fig.
\ref{figure:gp2}). 
The navigation maneuvers include lane following, three and four-way intersections, u-turns, and sharp
turns. The graph information is generated based on the closest
matching OSM waypoint with respect to the ego-vehicle using rough GPS
estimates and the positions of the nearby nodes are rotated based on
the orientation of the waypoint. This provides an OSM waypoint approximation without relying on an IMU. The dataset consists of $6,128$ training
and $2,864$ testing samples after interpolating groundtruth poses,
where each sample is composed of a global plan, a local semantic scene
representation, the groundtruth trajectory, IMU data, an Unix Epoch
timestamp, the state of the ego-vehicle in the semantic map frame
(within $2cm$ precision), and its state in a global frame (in terms of
latitude and longitude within $1m$ precision).

Secondly, a new dataset termed IntersectionScenes1.0 is introduced in
this work that focuses on evaluating the performance for three-way and
four-way intersection navigation. The dataset consists of $2,924$
training and $1,506$ testing samples after interpolation. For both
datasets, the global plans are generated by a GPS-based planner that
employs Dijkstra's shortest path search algorithm and generates
representations for the rasterized and graphical models. The graphical
version includes stop signs, traffic signals, and pedestrian crossings
as provided by OSM. For convenience, we retain the latitude and
longitude information from OSM to enable future planner
implementations.

\subsection{Metrics}
To evaluate the performance of each of our models, we measure the
quality of each trajectory generated in terms of driveable area
compliance (DAC) and trajectory quality. DAC measures the ability of the model to
generate trajectories that do not deviate from drivable areas; this
metric is computed by an average over all trajectories that overlap
with a drivable region, namely, crosswalks, lane marks, and road
surfaces as defined by the local semantic map. If any trajectory
waypoint overlaps with a region of sidewalk or vegetation, the prediction is strictly determined to be non-compliant. We report the error associated with half of the trajectory (DAC$_{HALF}$) and the complete predicted trajectory (DAC$_{FULL}$). Valid numerical values for compliance are given in the range $[0, 1]$.

Additionally, we leverage metrics that are commonly utilized to
benchmark the performance of trajectories in road user prediction
literature: Average Displacement Error (ADE) and Final Displacement
Error (FDE) \cite{fernando2018gdgan,gupta2018social}. By utilizing
these metrics, we can examine the average error for each trajectory
across all $H$ waypoints (ADE) and the average error associated with
the last waypoint predicted (FDE). An extension is performed to ADE by
measuring the error associated with half of the trajectory
(ADE$_{HALF}$) given that the waypoints closest to the autonomous
agent will be executed first during navigation. Finally, the worst
case errors are measured by the average maximum displacement error (MDE) along each predicted trajectory.

\begin{table*}[htbp]
  \vspace{2mm}
  \centering
  \begin{tabular}{||c c c c c c c||} 
    \hline
    Method & ADE$_{FULL}$ & ADE$_{HALF}$ & FDE & MDE & DAC$_{FULL}$ & DAC$_{HALF}$\\ [0.5ex] 
    \hline\hline
    Rasterized (baseline) & 1.056245 & $\mathbf{0.336941}$ & 2.447714  & 2.494614 & 0.849162 & 0.934218 \\ 
    Graph-STPF & $\mathbf{0.969206}$ & 0.353576 & $\mathbf{2.316740}$ & $\mathbf{2.393168}$ & $\mathbf{0.914869}$ & 0.944642 \\
    Graph-STCPF & 1.131581 & 0.388303 & 2.717636 & 2.795832 & 0.905864 & 0.942408 \\ 
    Graph-PF & 1.365685 & 0.538815 & 2.852894 & 3.047669 & 0.892321 & 0.933054\\ [1ex] 
    \hline
  \end{tabular}
  \caption{A comparison between multiple graph-based global plan encoders and a
    raster-based encoder (baseline) evaluated on the NominalScenes1.0 dataset. ADE, FDE, and MDE error is given in terms of meters. S denotes stop signs, T denotes traffic signals, C denotes
    pedestrian crossings, P denotes past trajectory, and F denotes
    future (planned) trajectory information. }
  \label{table:1}
\end{table*}

\begin{table*}[htbp]
  \centering
  \begin{tabular}{||c c c c c c c||} 
    \hline
    Method & ADE$_{FULL}$ & ADE$_{HALF}$ & FDE & MDE & DAC$_{FULL}$ & DAC$_{HALF}$\\ [0.5ex] 
    \hline\hline
    Rasterized (baseline) & 1.793062 & 0.673450 & 3.672231  & 3.728120 & 0.858367 & 0.913147\\  
    Graph-STPF & $\mathbf{1.511415}$ & $\mathbf{0.619056}$ & $\mathbf{3.087722}$ & $\mathbf{3.226302}$ & $\mathbf{0.898473}$ & $\mathbf{0.924834}$\\ [1ex] 
    \hline
  \end{tabular}
  \caption{Evaluation results for intersection navigation task (IntersectionScenes1.0 dataset). ADE, FDE, and MDE error is given in terms of meters. }
  \label{table:2}
\end{table*}

\subsection{Results}
To identify representative features for various modes of navigation,
three models are introduced and compared with respect to the
TridentNet CGM baseline~\cite{paz2021tridentnet} that utilizes
rasterized global plan representations. While all three of the studied
methods utilize a $(P+F)\times 3$ global plan representation, the
first one (Graph-PF) only incorporates the planned and traversed
trajectories, the second one (Graph-STCPF) additionally incorporates
stop signs, traffic signals, and pedestrian crossings and the third
model (Graph-STPF) only leverages stop signs and traffic signals in
addition to the planned and traversed trajectories.\footnote{Stop
  signs and traffic signals are encoded under the same category given
  that, in general, trajectories may not depend on stop sign vs
  traffic-light protected intersections.}

The results for each of the models evaluated with the NominalScenes1.0
test set are shown in Table \ref{table:1}. It can be observed that the model that embeds stop signs
and traffic signals only (STPF) outperforms the baseline and the
additional graph-based methods (STCPF and PF) in terms of
ADE$_{FULL}$, FDE, and MDE. On average, the first half of each
trajectory predicted (ADE$_{HALF}$) deviates by 35cm from groundtruth,
the complete trajectory composed of $H$ waypoints deviates by
approximately 97cm from groundtruth (ADE$_{FULL}$), the last
trajectory waypoint predicted deviates by 2.3m (FDE), and the worst
waypoint prediction deviates by 2.4m (MDE). In fact, we find that the
probability of a waypoint generating the worst error along a
trajectory is higher at the end point than any of the previous $H-1$
waypoints within a trajectory by 78\%.

\begin{figure}
\vspace{1.5mm}
\centering
  \begin{minipage}{.49\linewidth}
    \centering
    \subfloat[]{\label{figure:4a}\includegraphics[width=\textwidth]{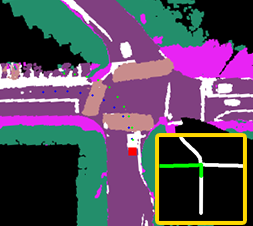}}
  \end{minipage}
  \hfill
  \begin{minipage}{.49\linewidth}
    \centering
    \subfloat[]{\label{figure:4b}\includegraphics[width=\textwidth]{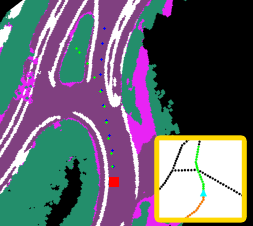}}
  \end{minipage}  
  
  \caption{Examples of DAC failure modes generated by the rasterized model (a) and the Graph-STPF model (b).}
  \label{figure:4}
\end{figure}

Although the rasterized version marginally outperforms Graph-STPF in
terms of ADE$_{HALF}$, this difference may be negligible and within
the localization error margin. The difference between STCPF and PF
implies that by incorporating stop signs, traffic signals, and
pedestrian crossings as additional node attributes, lower errors can
be achieved. These road features can be relevant for distinguishing
intersection navigation vs lane following. On the other hand, not all
road features improve the overall performance; this becomes evident
when comparing the model that uses pedestrian crossings (STCPF) and
the model that does not (STPF). We suspect that the performance
discrepancy occurs because crosswalk features are not exclusive to
intersection navigation. In contrast to stop signs and traffic
signals, crosswalks can be found during straight road segments and can
potentially introduce a bias towards intersections even though an
intersection does not exist.

Another interesting observation arises when evaluating the models with the DAC metrics. Although the error margin associated with half of each trajectory is minimal across most of the models, the graphical methods consistently outperform the baseline when we benchmark full trajectories. From our observations described in Section \ref{sec:global-plan-representation}, we find that the failure modes associated with the baseline are often due to IMU-based integration for heading estimation and ego-centric transformations. As a result, the predicted trajectories are more likely to deviate from the road if the heading is slightly off. In contrast, the graphical models use relative OSM waypoints to infer orientation and additionally incorporate information about the planned and traversed trajectory segments; thus preventing a bias towards IMU-based heading estimation. Examples with non-compliant trajectories can be visualized in Fig. \ref{figure:4}.

\begin{figure}[hbtp]
  
  
  \begin{minipage}{.32\linewidth}
    \centering
    \subfloat[]{\label{figure:5e}\includegraphics[width=\textwidth]{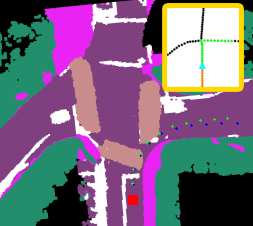}}
  \end{minipage}%
  \hfill
  \begin{minipage}{.32\linewidth}
    \centering
    \subfloat[]{\label{figure:5f}\includegraphics[width=\textwidth]{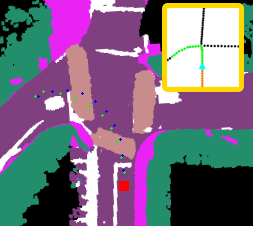}}
  \end{minipage}
  \hfill
  \begin{minipage}{.32\linewidth}
    \centering
    \subfloat[]{\label{figure:5g}\includegraphics[width=\textwidth]{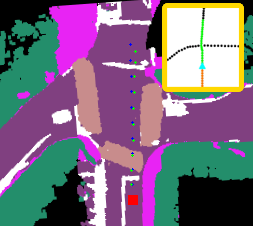}}
  \end{minipage}\par\medskip
  
  \caption{Test samples from the IntersectionScenes 1.0 dataset using the Graph-STPF approach. For each scenario, different global plans are provided to the model (left to right) while the ego-vehicle is at approximately the same location.}
\label{figure:5}
\end{figure}

To further explore the strengths of incorporating stop signs and
traffic signals, intersection navigation experiments are performed
using the IntersectionScenes1.0 dataset. The comparison is performed
specifically between STPF and the baseline model as shown in Table
\ref{table:2}. Although it is evident that the intersection navigation
task itself is inherently more challenging than lane following which
is included in Table \ref{table:1}, the results show consistently that
leveraging features that are intersection specific lead to performance
improvements and can aid in generating better trajectories while
entering and exiting an intersection. The multi-modal properties of
our generative model can be visualized in Fig. \ref{figure:5}, where
different global plans are defined from approximately the same
starting point.

\section{DISCUSSION}
The Self-Attention approach can encode a global plan
represented as a graph without depending on IMU-based heading
estimation. Compared to its rasterized counterpart that utilizes CNN
layers, improvements are achieved by incorporating road features
readily available from OSM. Moreover, by introducing graphical data
instead of image-based tensors, the number of parameters of the model
is drastically reduced by 31\% as shown in Table \ref{table:3}.

An additional consideration for robotics applications includes the
real-time capabilities of the model. Given that our pose and
perception information is updated at a rate of 10Hz, the processing time for the autonomy stack should
remain below 100ms. All models evaluated in average have an inference time of
approximately 6ms, implying that only 6\% of the allotted time is
utilized. Thus, the real-time characteristics of the model can provide
additional flexibility for perception and decision-making modules that
may require more computational time.

\begin{table}[hbtp]
  \begin{center}
    \begin{tabular}{|c||c||c|}
      \hline
      Model & Model Parameters & Avg. Inference Time \\
      \hline
      w/ Self-Attention & $\mathbf{11.5}$M & 6.18ms\\
      \hline
      w/ CNN Encoder & 16.8M & 6.22ms\\
      \hline
    \end{tabular}
    \caption{Size of the model and associated inference times.}
    \label{table:3}
  \end{center}
\end{table}

Additional experiments were performed with Graph Convolutional
Networks \cite{kipf2017semisupervised} and LSTM-based encoders;
however, performance improvements were not observed with respect to
the Self-Attention model. While these results are not shown in our
tables, these methods could be useful to further explore the
spatial-temporal relationship as a graph by embedding complete road
information over a sliding window. Nevertheless, these additional
experiments demonstrate that our proposed OSM graphical representation
can be applied with similar learning-based encoders. 

\section{CONCLUSION}

In this work, graph-based representations for global plans are
studied in the context of dynamic trajectory generation without
relying on HD maps. The results show that by leveraging a
Self-Attention mechanism, low relative errors can be achieved with
respect to the baseline composed of rasterized global plan
representations whilst simultaneously reducing the complexity of
generating and operating with image representations.

For future work, we plan to investigate robust methodologies for
incorporating physical constraints into our generative model, apply
our method as a shared mechanism for road user behavior prediction,
and further investigate how to leverage real-time scene
representations for dynamic trajectory generation.


\addtolength{\textheight}{-6cm}   





\section*{ACKNOWLEDGMENT}
We appreciate the feedback and comments provided by members of the
Autonomous Vehicle Lab. 
We also thank the University of California, San Diego for providing 
campus access for the data collection process
completed in this study.

\newpage

%

\bibliographystyle{./IEEEtran}
\IEEEtriggeratref{13} 
\bibliography{IEEEcitation}

\end{document}